\documentclass{article} 
\usepackage{iclr2017_workshop,times}
\usepackage{hyperref}
\usepackage{url}
\usepackage{graphicx}
\usepackage{amsmath}

\title{Unsupervised Feature Learning for Audio Analysis}

\author{Matthias Meyer, Jan Beutel, Lothar Thiele \\
Computer Engineering and Networks Laboratory\\
ETH Zurich\\
Zurich, Switzerland \\
\texttt{\{matthias.meyer, beutel, thiele\}@tik.ee.ethz.ch} \\
}

\begin{document}

\maketitle

\begin{abstract}
Identifying acoustic events from a continuously streaming audio source is of interest for many applications including environmental monitoring for basic research. In this scenario neither different event classes are known nor what distinguishes one class from another. Therefore, an unsupervised feature learning method for exploration of audio data is presented in this paper. It incorporates the two following novel contributions: First, an audio frame predictor based on a Convolutional LSTM autoencoder is demonstrated, which is used for unsupervised feature extraction. Second, a training method for autoencoders is presented, which leads to distinct features by amplifying event similarities. In comparison to standard approaches, the features extracted from the audio frame predictor trained with the novel approach show 13 \% better results when used with a classifier and 36 \% better results when used for clustering.
\end{abstract}

\section{Introduction}
The advance of low-power mobile sensors leads to new applications for real-time environmental monitoring, which include acoustic event detection (AED) systems using wireless sensor networks, e.g. for vehicle classification (\cite{duarte_vehicle_2004}). Acoustic events can be effectively classified by convolutional neural networks (CNN) as shown by \cite{espi_exploiting_2015}, if a large labeled training dataset is available (\cite{hershey_cnn_2016}). However, for some application scenarios, e.g. in micro-seismic acoustic emission monitoring
(\cite{girard_custom_2012}), the advantages of mentioned CNNs are diminished due to the lack of ground truth. In this scenario distinguishing between relevant events in an audio stream and non-relevant events is non-trivial, since the domain-specific categories of these acoustic events are not known a priori. Unsupervised feature learning provides a mean to gather useful information from such unexplored datasets, which has been shown for video (\cite{srivastava_unsupervised_2015}) and audio (\cite{honglak_lee_unsupervised_2009}) analysis. Recently, convolutional long short-term memory (ConvLSTM) layers have proven to be effective as feature extractors for time series data (\cite{xingjian_convolutional_2015}) and for unsupervised video analysis (\cite{finn_unsupervised_2016}). Except for speech recognition (\cite{zhang_very_2016}), ConvLSTM layers have not been used for audio analysis, despite their advantages for time series data. 

As a consequence, in this paper a novel audio frame predictor (AFP) using ConvLSTM layers is presented, which is used to learn representative features of acoustic events. Second a novel approach to train the AFP in regard to generating and extracting distinct features is presented which diversifies the features based on inter-sample similarities.

\begin{figure}[h]
	\begin{center}
		\includegraphics[width=10cm]{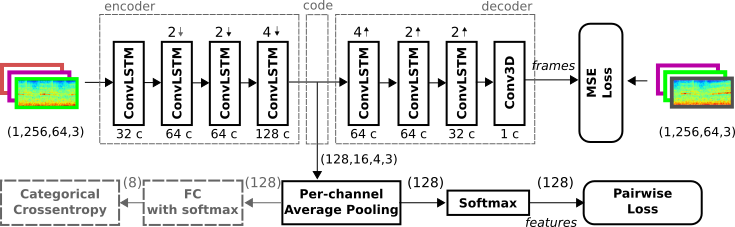}
	\end{center}
	\caption{Model of the audio frame predictor, which is trained to predict the next frame of the mel-spectrogram input. Multiple ConvLSTM layers are used which differ by number of channels (e.g. 32 c) and down-/upsampling (e.g. $2\downarrow$ ). 
	Per-channel average pooling reduces the code to a lower dimensional feature vector on which a pairwise loss is applied. For evaluation a simple classifier, consisting of a fully connected (FC) layer, classifies events based on the extracted features.}
	\label{fig_model}
\end{figure}

\section{Approach}
An autoencoder encodes an arbitrary input signal into a compressed intermediate representation (code) from which it tries to recreate the original signal (\cite{hinton_reducing_2006}).
In this section it is assumed that an autoencoder using convolutional layers, as depicted in Figure \ref{fig_model}, can be trained such that each channel of the code contains information about a specific feature of acoustic events. Then, also the mean of this channel would define an acoustic feature and per-channel average pooling can be used to reduce the $K$-channel code to a $K$-dimensional vector (feature vector). A softmax function transforms this vector into a categorical distribution (from here on called feature distribution)

In this work, similarities between two audio samples are characterized by the statistical distance between their feature distributions, a concept similar to \cite{hsu_neural_2015}. Since training of a neural network is typically done in batches of size $N$, a pairwise loss function $L(n,m)$ for the \emph{features} output can be defined as:


\begin{equation}
L(n,m) = 
\begin{cases}
\mathit{KL}(B^{(n)} \mid\mid B^{(m)}), & \text{if } f_{n,m} < \text{threshold} \\
\max( 0, \mathit{margin} -KL(B^{(n)} \mid\mid B^{(m)})), & \text{else}
\end{cases}
\end{equation}

\begin{equation}
f_{ n,m } = \frac{{KL(B^{(n)} \mid\mid B^{(m)})}}  {\frac{1}{N^2}{\sum_{o=0}^{N-1} \sum_{p=0}^{N-1} KL(B^{(o)} \mid\mid B^{(p)})}} 
\end{equation}

where $B^{(n)}$ defines the feature distribution of the n-th element of the batch. $KL$ is the Kullback-Leibler divergence and the two parameters \emph{threshold} and \emph{margin} can be adjusted based on the actual data. $f_{n,m}$ is a measure for the similarity of two elements of the batch. Therefore thresholding $f_{n,m}$ means that the loss $L(n,m)$ penalizes events based on similarity, for similar events the feature distribution gets more confined while dissimilar events are shifted away from each other. Thus, no (weak) labels are required in contrast to \cite{hsu_neural_2015}. The overall loss per batch is $\sum_{n=0}^{N-1} \sum_{m=0}^{N-1} L(n,m)$, which can be backpropagated through the network and affects the weights of the encoder. The total loss for the encoder weights is the sum of the autoencoder loss and the pairwise loss. The autoencoder enforces a meaningful representation while $L(n,m)$ amplifies similarities and dissimilarities. Thus, this method optimizes for inter-sample similarities in comparison to the method from \cite{xie_unsupervised_2016}, which optimizes for fixed cluster centers.

For the actual implementation an autoencoder is used as an AFP, which is illustrated in Figure \ref{fig_model}. Data frames of length $T_f=2.56\text{ s}$ are extracted from the audio waveform with a hopsize of $T_f/4$ and transformed into a mel-spectrogram. The AFP considers three consecutive frames and predicts the following frame.
Each ConvLSTM layer uses a 3x3 filter kernel and is followed by a ReLu activation and batch normalization. The last convolutional layer (3D convolution) applies a 3x3x1 kernel with a linear activation function. Keras (\cite{chollet_keras_2015}) and Tensorflow (\cite{martin_abadi_tensorflow:_2015}) are used for the implementation.

The training with batchsize $N=16$ is done in two steps. First, only the AFP is trained by minimizing the mean squared error (MSE) between the \emph{frames} output of the AFP and the true next frame, initializing the weights of the whole system. After training the AFP for 6 epochs the system is trained for another 3 epochs by minimizing the above mentioned pairwise loss at the \emph{features} output and the MSE at the \emph{frames} output, simultaneously. In the next section a system trained with this two-step procedure (PairLoss) will be compared against a system trained only by minimizing the MSE at the \emph{frames} output for 9 epochs (w/o PairLoss).

\section{Evaluation}
\begin{figure}[h]
	\begin{center}
		\includegraphics[width=11.5cm]{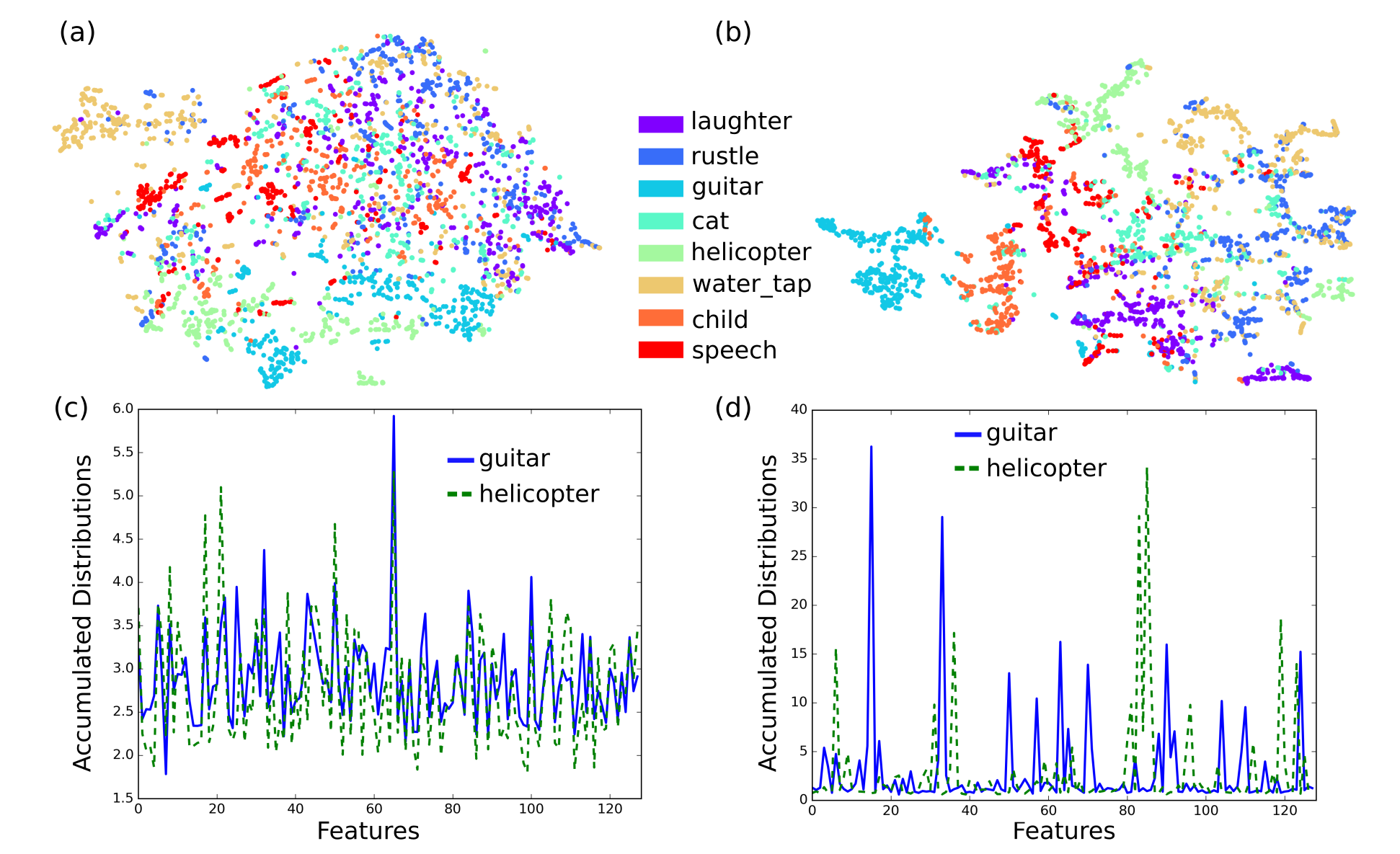}
	\end{center}
	\caption{t-SNE embedding (\cite{maaten_visualizing_2008}) of the feature distributions for a system trained without PairLoss (a) and with PairLoss (b). Accumulated feature distributions for all training samples of the category \emph{guitar} and \emph{helicopter} for a system trained without PairLoss (c) and with PairLoss (d). Interactive version at: http://people.ee.ethz.ch/matthmey/  }
	\label{fig_tsne}
\end{figure}

An acoustic event dataset (AED) from \cite{takahashi_deep_2016} was used for evaluation. In Figure \ref{fig_tsne} eight categories are listed which were chosen to have an approximate uniformly distributed number of samples. The dataset has been partitioned into training (75\%) and test set (25\%). The number of code channels has been set to $K=128$ to maintain an accurate prediction and to provide enough features to explain the data.

\begin{table}[h]
\caption{Test accuracies for classifier and clustering algorithm. Note: The reference classifier uses supervised training for all weights, the others only for the fully connected layer}
\label{tab_acc}
\begin{center}
\begin{tabular}{lll|ll}
\multicolumn{3}{c|}{\bf Classifier} & \multicolumn{2}{c}{\bf Clustering} \\
Reference & w/o PairLoss & w/ PairLoss & w/o PairLoss & w/ PairLoss
\\ \hline
(90.85 \%) & 69.59 \% & \bf 78.94  \% & 38.32 \% & \bf 52.23 \% \\

\end{tabular}
\end{center}
\end{table}

First, it is shown that the feature vectors contain useful information by learning a mapping from the 128-dimensonal feature vectors to the 8 categories. A simple classifier, consisting of only one fully connected layer, as depicted in Figure \ref{fig_model}, is trained by minimizing the cross-entropy loss between the classifier's output and the ground-truth labels. The weights of the encoder are fixed. The PairLoss system outperforms the one without PairLoss on the test set, which can be attributed to the different feature distributions, as illustrated in Figure \ref{fig_tsne} (c) and (d). The plot displays the accumulated feature distributions for two event categories. In contrast to standard training, it becomes apparent that for PairLoss training two different event categories show distinct patterns for their feature distributions, which means that only a few channels contain event information. It is also shown that using a system trained with PairLoss shows better clustering accuracies than a system trained without, as presented in Table \ref{tab_acc}. Although the system can potentially be expanded to perform clustering itself, for this evaluation only k-means clustering (8 clusters) is applied to the t-SNE embedding of the feature distributions. The best cluster assignment is chosen using the Hungarian algorithm (\cite{kuhn_hungarian_1955}).

In this work a novel approach to train autoencoders has been used in a novel unsupervised feature extraction system for audio data. It has been shown experimentally that training with the new method generates distinct features which increases classification accuracy by 13 \% and clustering accuracy by 36 \%.



\bibliography{iclr2017_workshop}
\bibliographystyle{iclr2017_workshop}

\end{document}